\documentclass[runningheads]{llncs}
\usepackage[T1]{fontenc}
\usepackage{graphicx}
\usepackage{marvosym}

\usepackage{amssymb}
\usepackage{subcaption}
\usepackage{booktabs}
\usepackage{multirow}
\usepackage{graphicx}
\usepackage[table]{xcolor}
\usepackage{xcolor}
\usepackage{pifont}
\definecolor{mygrey}{HTML}{999999}
\definecolor{myblue}{HTML}{336699}
\definecolor{mygreen}{HTML}{669933}
\usepackage{tikz}
\usepackage{hyperref}
\usepackage{amsmath}

\begin{document}

\title{AgriVLN: Vision-and-Language Navigation \\for Agricultural Robots}

\titlerunning{AgriVLN}

\author{
Xiaobei Zhao\textsuperscript{\rm 1,2}\textsuperscript{(\Letter)} \and
Xingqi Lyu\textsuperscript{\rm 1} \and
Xin Chen\textsuperscript{\rm 1,2}\textsuperscript{(\Letter)} \and
Xiang Li\textsuperscript{\rm 1,2}\textsuperscript{(\Letter)}
}

\authorrunning{Accepted as a conference paper at ICONIP 2026}

\institute{\textsuperscript{\rm 1}China Agricultural University\\
\textsuperscript{\rm 2}China Agricultural University-Sichuan Advanced Agricultural \& Industrial Institute\\
\email{\{xiaobeizhao2002,lxq99725\}@163.com, \{chxin,cqlixiang\}@cau.edu.cn}}


\maketitle              
%


\begin{figure}[h]
\vspace{-5pt}
\centering
\includegraphics[width=1.0\columnwidth]{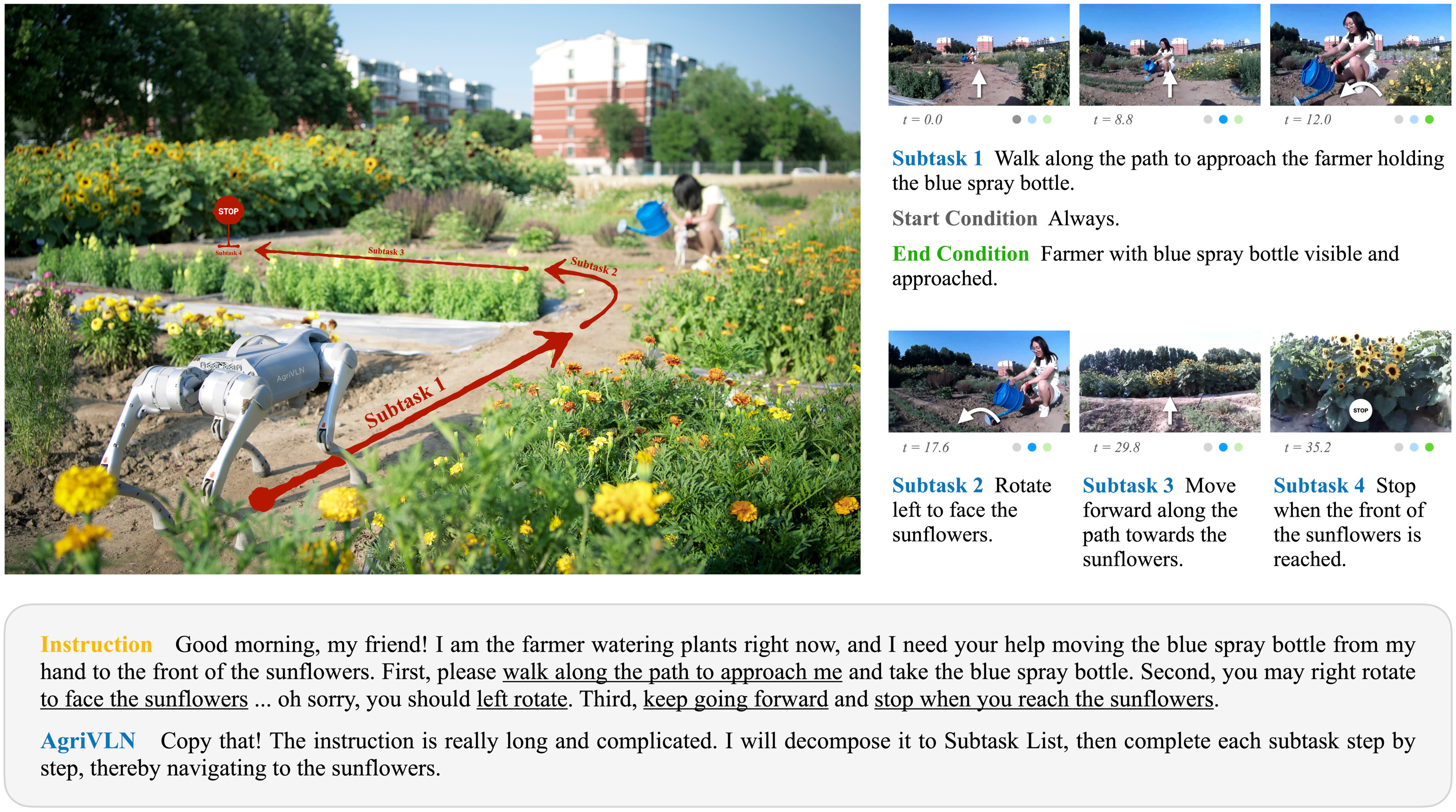}
\caption{AgriVLN illustration on a representative episode. After receiving an instruction from user, AgriVLN decomposes it into a list of subtasks, then executes each subtask one by one, thereby reasoning a sequence of low-level actions to navigate an agricultural robot from a starting point to a target position.}
\label{fig:teaser}
\vspace{-20pt}
\end{figure}

\begin{abstract}
  
Agricultural robots are serving as powerful members across a wide range of tasks, nevertheless, they are still heavily relying on manual operations or railway systems for movement. In contrast, Vision-and-Language Navigation (VLN) enables navigating robots from starting points to target positions following natural language instructions, having shown strong performances across various domains. However, none of the existing benchmarks or methods are specifically designed for agricultural robots. To bridge this gap, we build our A2A benchmark, which consists of 1,560 episodes across 6 classes of agricultural scenes using the step-by-step format for instructions and the continuous environment for actions. We evaluate several existing state-of-the-art VLN methods on A2A, then we observe that they perform well on short instructions while perform poorly on long instructions. To alleviate this phenomenon, we design our STL module, which decomposes a complex instruction into a list of simple subtasks, thereby assisting the decision-maker to focus on the current short subtask rather than the entire long instruction at any time. We integrate STL into the NavGPT backbone to propose our AgriVLN method, achieving a sufficient improvement on SR from 0.33 to 0.47 with a slight sacrifice on NE from 2.76 m to 2.91 m. We further implement the ablation studies and the case study, discussing the effectiveness and limitations of AgriVLN on different subtask quantities and scene classes. Repository: \href{https://github.com/AlexTraveling/AgriVLN}{https://github.com/AlexTraveling/AgriVLN}.


\end{abstract}

\section{Introduction}

Agricultural robots are serving as powerful members across a wide range of agricultural tasks, such as laser weeding \cite{COMPAG:LaserWeeding}, cross-pollination \cite{Cell:Pollination} and phenotypic measurement \cite{ICIC:TomatoScanner}. However, most of them are still heavily relying on manual operations or railway systems for movement, which results in limited mobility and poor adaptability \cite{COMPAG:Revealing,ASOC,IJCAI:Survey}. 

\par In contrast, Vision-and-Language Navigation (VLN) \cite{CVPR:R2R,ECCV:VLN-CE} enables navigating a robot from a starting point to a target position following a natural language instruction, having demonstrated strong performance across various domains \cite{TMLR}, such as R2R \cite{CVPR:R2R} for indoor rooms, TOUCHDOWN \cite{CVPR:TOUCHDOWN} for urban streets, and AerialVLN \cite{ICCV:AerialVLN} for air spaces. However, none of the existing benchmarks, as far as we know, are specifically designed for agricultural robots, from which we suggest a needed yet lacked benchmark for training and evaluating VLN methods for agricultural robots. 

\par In this paper, we build the benchmark of \textbf{A}griculture-\textbf{to}-\textbf{A}griculture (\textbf{A2A}). We collect a total of 1,560 episodes with an average instruction length of 45.5 across 6 classes of agricultural scenes, including farms, greenhouses, forests, mountains, gardens and villages, basically supporting all the common scenarios. We design the instructions in the step-by-step format and the actions in the continuous environment. 

\par We reproduce several existing state-of-the-art VLN methods on our A2A benchmark, then we observe that they perform well on short instructions while perform poorly on long instructions. We attribute this interesting phenomenon to the challenging semantic alignment: Their decision-makers often fail to track which part of the instruction is currently being executed.

\par To alleviate this phenomenon, we get inspiration from a daily life habit: The to-do list. When met a complex job, a person might have the subconscious to decompose it into a list of to-do items, then did them one by one, thereby helping itself better focus on one item at any time. Inspired by this habit, we design the module of \textbf{S}ub\textbf{t}ask \textbf{L}ist (\textbf{STL}). It decomposes a complex instruction into a list of simple subtasks, thereby assisting the decision-maker to focus on the current short subtask rather than the entire long instruction at any time.

\par We integrate it into the NavGPT \cite{AAAI:NavGPT} backbone to propose the method of \textbf{V}ision-and-\textbf{L}anguage \textbf{N}avigation for \textbf{Agri}cultural Robots (\textbf{AgriVLN}). After receiving an instruction from user, our AgriVLN method decomposes it into a list of subtasks, then executes each subtask one by one, thereby reasoning a sequence of low-level actions to navigate an agricultural robot from a starting point to a target position. When evaluated on A2A, we observe a sufficient improvement on Success Rate (SR) from 0.33 to 0.47 with a slight sacrifice on Navigation Error (NE) from 2.76 m to 2.91 m, from which we suggest that our STL module can effectively alleviate the challenging semantic alignment. 

\par Here we share a simple example to better demonstrate how AgriVLN runs on A2A, as illustrated in Fig. \ref{fig:teaser}. After receiving the instruction, AgriVLN decomposes it into a list of four subtasks, then navigate the agricultural robot to move $\texttt{[FORWARD]}$ to reach the user, $\texttt{[LEFT ROTATE]}$ to face the target sunflowers, move $\texttt{[FORWARD]}$ to approach them, and $\texttt{[STOP]}$ when reaching them.

\par In summary, our main contributions are as follows:
\begin{itemize}
\item \textbf{A2A}, the first VLN benchmark for agricultural robots, which consists of 1,560 episodes across 6 classes of agricultural scenes using the step-by-step format for instructions and the continuous environment for actions. 
\item \textbf{STL}, a plug-and-play module, which decomposes an instruction into a list of subtasks, thereby assisting the decision-maker to focus on the current short subtask rather than the entire long instruction at any time. 
\item \textbf{AgriVLN}, the first VLN method for agricultural robots, which enables navigating an agricultural robot from a starting point to a target position following a natural language instruction. 
\end{itemize}

\section{Related Works}

\subsection{Benchmark of VLN}

In addition to the traditional VLN benchmarks in residences, several recent studies have extended VLN to more scenes. TOUCHDOWN \cite{CVPR:TOUCHDOWN} collected the street views in New York City, supporting the evaluation for urban service robots. AerialVLN \cite{ICCV:AerialVLN} and UAV-Need-Help \cite{ICLR:UAV-Need-Help} collected the aerial photos on different terrains, supporting the evaluation for unmanned aerial vehicles. Nevertheless, none of the existing benchmarks contain images from agricultural scenes.

\subsection{Vision-Language Model for VLN}

Traditional VLN methods \cite{NeurIPS:HAMT,NeurIPS:Landmark,NeurIPS:ZSON} usually used Reinforcement Learning (RL) \cite{TETCI:Efficient} for decision-making, which performed well in seen environments but struggled with unseen environments, showing the poor generalization capability. To address this weakness, several recent studies \cite{AAAI:NavGPT,ICRA:Pixel,ACL:MapGPT} introduced Vision-Language Models (VLMs) for decision-making, thereby achieving the zero-shot reasoning with better generalization capability.

\subsection{Instruction Decomposition in VLN}

\par Most of previous VLN methods \cite{ICCV:YouTube,AAAI:VELMA,CVPR:OVER-NAV} directly took a whole instruction as an input for decision-making. As the instruction becomes more and more complex, it is more and more difficult for a decision-maker to align a linguistic description with a visual object \cite{AAAI:VLN-VIDEO,ICCV:VLN-PETL,ACL:BabyWalk}. To address this weakness, several recent studies \cite{EMNLP:SIA-VLN,RAL:DILLM-VLN} explored to chunk an instruction into a list of sub-instructions then execute every sub-instruction in order, effectively assisting the decision-maker to only focus on a small part of the instruction at each time step.

\section{Benchmark}

In this section, we present the A2A benchmark. First, we introduce the task definition in Sec. \ref{sec:task_definition}. Second, we present the dataset collection in Sec. \ref{sec:dataset_collection}. Third, we present the dataset assessment in Sec. \ref{sec:dataset_assessment}. Fourth, we introduce the evaluation metrics in Sec. \ref{sec:evaluation_metrics}.

\subsection{Task Definition}
\label{sec:task_definition}

We define the task of Agricultural Vision-and-Language Navigation as follows: In every episode, the model is given an instruction in natural language, denoted as $W = \langle w_1, w_2, \dots, w_L \rangle$, where $L$ is the number of words. At each time step $t$, the model is given a front-facing RGB image $I_t$. The purpose is understanding both $W$ and $I_t$ to predict the most appropriate low-level action $\hat{a_t}$ from the action space $\{ \texttt{[FORWARD]}$, $\texttt{[LEFT ROTATE]}$, $\texttt{[RIGHT ROTATE]}$, $\texttt{[STOP]} \}$, thereby navigating the agricultural robot from the starting point to the target position.

\subsection{Dataset Collection}
\label{sec:dataset_collection}

To support diversified classes of agricultural scenes, we organize the dataset collection at three locations in Beijing: China Agricultural University, China Agricultural University Shangzhuang Test Station, and Olympic Forest Park.

\subsubsection{Instruction Generation} 

In every episode, according to a practical agricultural activity, our experts design a step-by-step instruction to describe the moving trajectory of the worker. Different from traditional VLN benchmarks' delicate and concise instructions, we design A2A's instructions in a more casual and tedious style with a lot of nonsense and misleading noises, thereby better aligning with the practical speaking styles of the agricultural users.

\subsubsection{Robot Control} 

We adopt the Unitree Go2 Air four-legged robotic dog as the experimental agricultural robot. In every episode, our experts manually control it moving on the most appropriate route from the starting point to the target position, thereby succeeding the corresponding instruction.

\subsubsection{Video Recording}

We adopt the built-in front-facing RGB camera of the Unitree Go2 Air four-legged robotic dog as the video recorder. In every episode, our experts manually operate it to record the front-facing views across all the time steps. Every video streaming is recorded at the 1280 $\times$ 720 resolution with a frame rate of $\sim$14 FPS and a bitrate of $\sim$1100 kbps.

\subsubsection{Data Annotation} 

For each time step $t$, our experts manually annotate the ground-truth action $a_t$ according to the practical moving state. We cluster a sequence of adjacent identical actions $\langle a_{t_1}, a_{t_1+1}, a_{t_1+2}, \dots, a_{t_2} \rangle$ into a temporal interval in the dictionary format $\{$action: $a_{t1}$, time range: $[t_1, t_2]\}$. Every episode consists of several temporal intervals, being saved in the JSON format.

\subsection{Dataset Assessment}
\label{sec:dataset_assessment}

\begin{figure}[t]
\centering
\includegraphics[width=0.8\columnwidth]{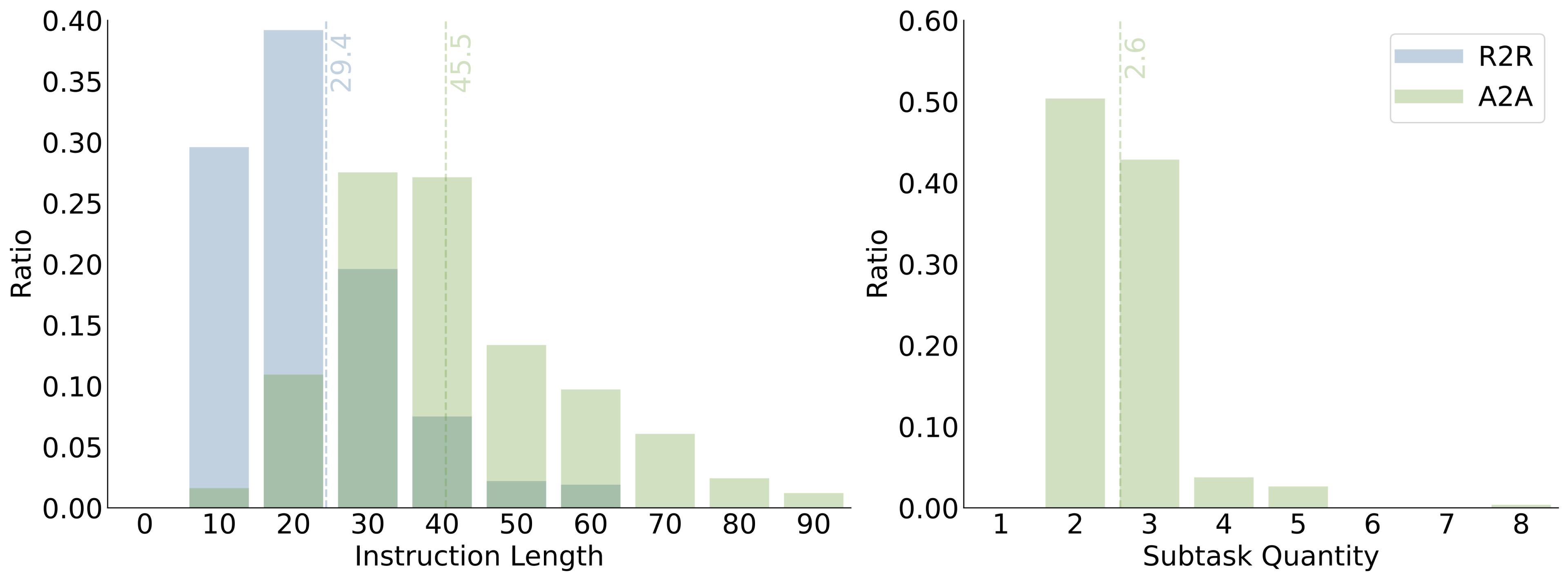}
\caption{A2A distribution illustration on instruction lengths and subtask quantities. The dotted lines represent the average values.}
\label{fig:benchmark_distribution}
\end{figure}

\begin{figure}[!t]
  \centering
  \begin{subfigure}[b]{0.4\columnwidth}
    \includegraphics[width=\linewidth]{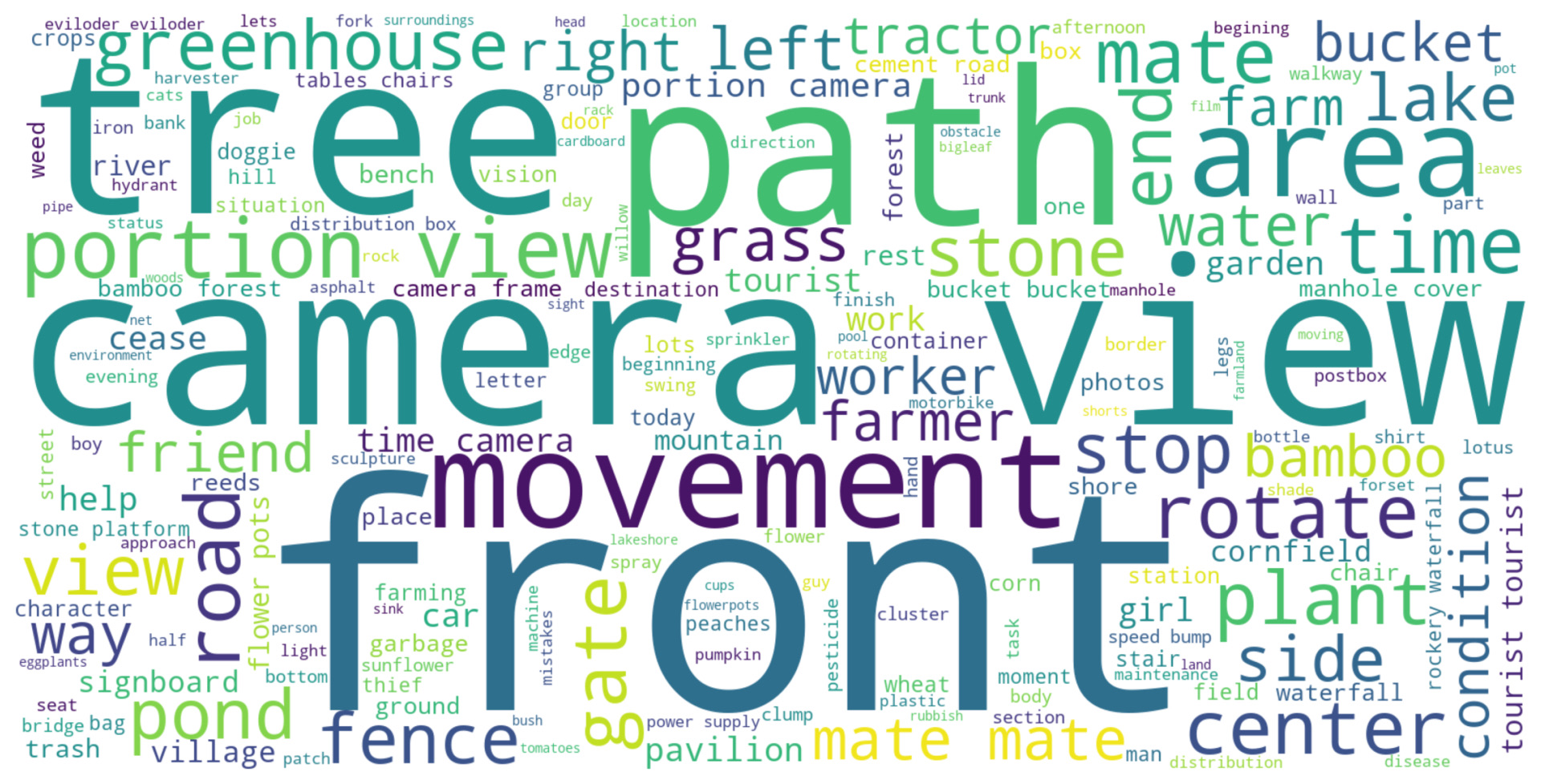}
    \caption{Nouns.}
  \end{subfigure}
  \begin{subfigure}[b]{0.4\columnwidth}
    \includegraphics[width=\linewidth]{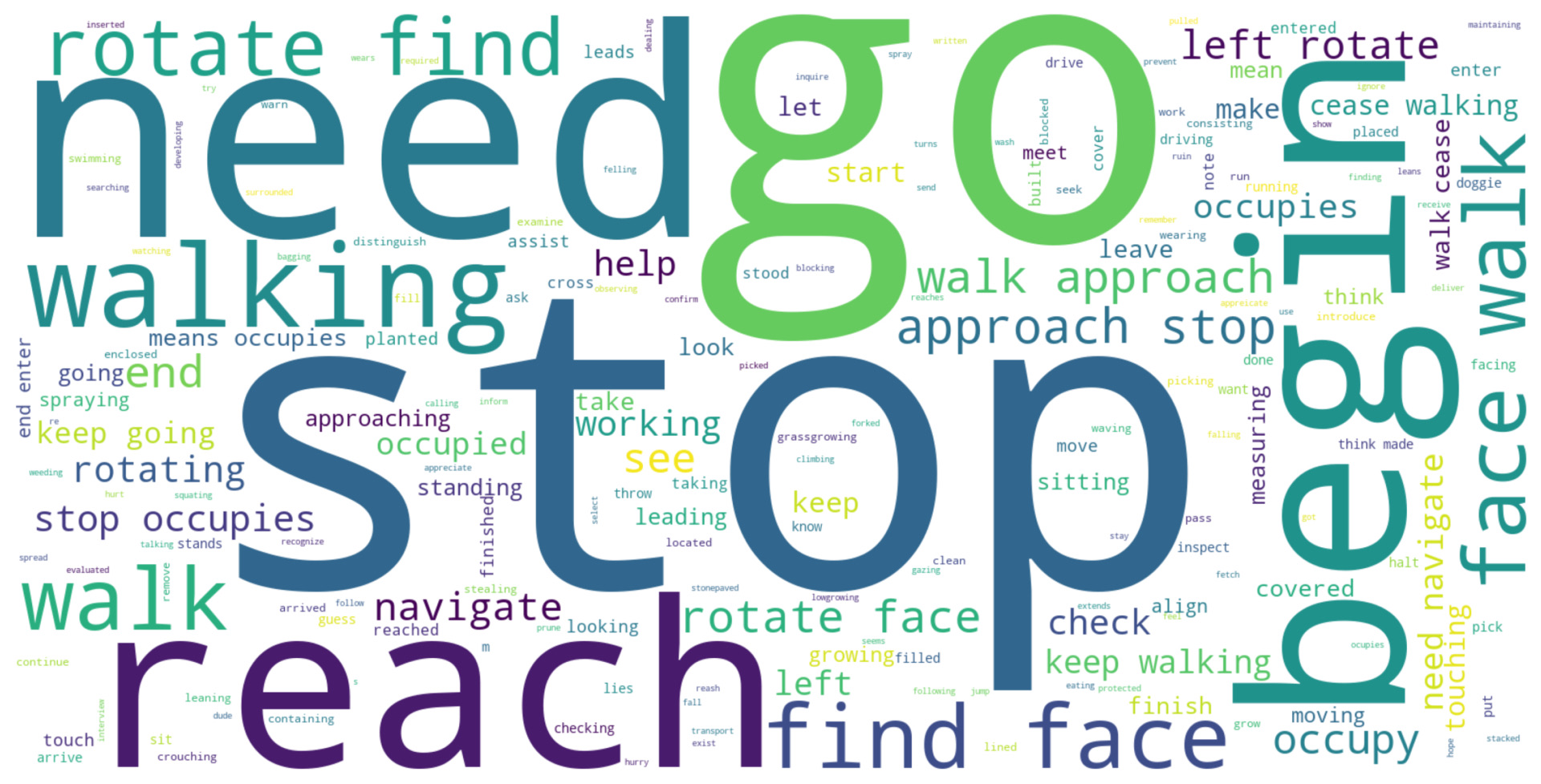}
    \caption{Verbs.}
  \end{subfigure}
  \caption{A2A wordcloud illustration on different classes of words. The larger the font size, the higher the appearance frequency.}
  \label{fig:wordcloud}
\end{figure}

In our A2A benchmark, we collect a total of 1,560 episodes across 6 classes of scenes, including 372 in farms, 258 in greenhouses, 384 in forests, 198 in mountains, 258 in gardens, and 90 in villages. In terms of weather, about 70\% are collected under sunny conditions and about 30\% are collected under overcast conditions. In terms of illumination, about 95\% are recorded during daytime and about 5\% are recorded during nighttime. As illustrated in Fig. \ref{fig:benchmark_distribution}, the lengths of the instructions range from 10 to 99 with an average length of 45.5, and the numbers of the subtasks range from 2 to 8 with an average number of 2.6. As illustrated in Fig. \ref{fig:wordcloud}, a total of 893 words appears in the vocabulary, in which “front”, “camera” and “view” are commonly utilized as nouns, and “go”, “stop” and “need” are commonly utilized as verbs. In summary, all the words we select are common in daily life, showing the alignment on speech tone between the instructions within A2A and the conversations within agricultural users.

\par We additionally compare our A2A benchmark with several mainstream VLN benchmarks from different aspects, including R2R \cite{CVPR:R2R}, REVERIE \cite{CVPR:REVERIE}, TOUCHDOWN \cite{CVPR:TOUCHDOWN}, VLN-CE \cite{ECCV:VLN-CE} and AerialVLN \cite{ICCV:AerialVLN}, as shown in Table \ref{tab:Benchmark Comparison}.

\subsubsection{Scene Diversity} 

Traditional VLN benchmarks \cite{CVPR:R2R,CVPR:REVERIE,ECCV:VLN-CE} mainly focus on indoor scenes, therefore, almost all the data are collected in residences. However, there is still a large gap between different indoor scenes, e.g., a robotic vacuum cleaner may work well in an organized kitchen but work bad in a chaotic warehouse. Hence, we suggest that their limited scene diversities cannot fully evaluate the generalization capability. Compared to them, AerialVLN \cite{ICCV:AerialVLN} collects data from 25 classes of aerial scenes, such as downtowns, factories and parks, thereby supporting the scene diversity. In our A2A benchmark, we collect data from 6 classes of agricultural scenes, thereby supporting all the common agricultural scenarios, from which we suggest the excellent scene diversity to our A2A.

\begin{table*}[t]
\captionsetup{skip=8pt}
\caption{VLN benchmark assessment results. “Cls.”, “Inst.” and “Traj.” denote class number, instruction length and trajectory length, respectively. “I.”, “O.”, “P.”, “S.”, “T.” and “E.” denote indoor, outdoor, panoramic, single, training and evaluation, respectively. Units of “Height” and “Traj.”: meter.}
\label{tab:Benchmark Comparison}
\centering
\resizebox{\linewidth}{!}{
\renewcommand{\arraystretch}{1.3}
\setlength{\tabcolsep}{2pt}
\begin{tabular}{lcc|ccc|ccc}
\toprule
\multirow{2}{*}{\textbf{Benchmark}} &
\multicolumn{2}{c}{\textbf{Scene}} & \multicolumn{3}{c}{\textbf{Camera}} & \multicolumn{3}{c}{\textbf{Scale}} \\
\cmidrule(lr){2-3} \cmidrule(lr){4-6} \cmidrule(lr){7-9} 
& \textbf{Domain} & \textbf{Cls.} & \textbf{Image} & \textbf{Height} & \textbf{Real} & \textbf{Inst.} & \textbf{Traj.} & \textbf{Episode} \\
\midrule
R2R        & Residence (I.) & 1  & RGB-D (P.) & 1.5     & $\surd$  & 29 & 10.0 & 7,189 (T.\&E.) \\
REVERIE    & Residence (I.) & 1  & RGB-D (P.) & 1.5     & $\surd$  & 18 & 10.0 & 21,702 (T.\&E.) \\
TOUCHDOWN  & Street (O.)    & 1  & RGB (P.)   & 2.5     & $\surd$  & 90 & 313.9 & 9,326 (T.\&E.) \\
VLN-CE     & Residence (I.) & 1  & RGB-D (S.) & 1.5     & $\surd$  & 19 & 11.1 & 4,475 (T.\&E.) \\
AerialVLN  & Air (O.)       & 25 & RGB-D (S.) & $>$ 5.0 & $\times$ & 83 & 661.8 & 8,446 (T.\&E.) \\
\rowcolor{gray!15}
A2A (Ours) & Agri. (I.\&O.) & 6  & RGB (S.)   & 0.38    & $\surd$  & 46 & 7.1 & 1,560 (E.) \\
\bottomrule
\end{tabular}
}
\end{table*}

\subsubsection{Image Quality} 

For a VLN benchmark, the image class should align with the practical applications. Take REVERIE \cite{CVPR:REVERIE} as an example: Panoramic RGB-D images are costly and demanding, but can be provided by a robotic vacuum cleaner. However, most of agricultural robots only have one front-facing RGB camera. Meanwhile, most of VLN benchmarks capture images at a relatively high height, such as TOUCHDOWN \cite{CVPR:TOUCHDOWN} equips the camera on a car at a height of 2.5 m, while most of agricultural robots are designed in small sizes. In addition, AerialVLN \cite{ICCV:AerialVLN} collects synthetic images from Unreal Engine 4 and Microsoft AirSim plugins, which improves the collection efficiency, but brings the sim-to-real gap. Hence, we suggest that A2A capturing RGB images at a height of 0.38 m in reality completely aligns with the practical agricultural applications.

\subsubsection{Data Scale} 

In terms of instruction length, several VLN benchmarks only provide short instructions, such as the average instruction length of R2R \cite{CVPR:R2R} is only 29.4, which cannot satisfy users to describe a complex task. Compared to them, our A2A surpasses on task complexities with an average instruction length of 45.5, which supports evaluating the performances on long instructions. In terms of episode number, our A2A is relatively less than others, while we design AgriVLN running in a training-free manner (will be presented in Sec. \ref{sec:methodology}), then all the data are available for evaluation. Hence, we suggest that the data scale of our A2A is sufficient for evaluating agricultural VLN methods.

\subsection{Evaluation Metrics}
\label{sec:evaluation_metrics}

We follow the two standard VLN evaluation metrics \cite{CVPR:R2R}: Success Rate (SR) and Navigation Error (NE). NE measures the path length between the stopping position and the target position. SR measures the rate successfully reaching the target position within a 2-m NE. Besides, we introduce Independent Success Rate (ISR) from LH-VLN \cite{CVPR:LH-VLN} to evaluate the success rate of each subtask individually, calculated as: 
\begin{equation}
ISR = \frac{\mathbb{E}_{i \in \mathcal{E}} \left[ SQ_i \right]}{\mathbb{E}_{i \in \mathcal{E}} \left[ TQ_i \right]}
\end{equation}
where $SQ_i$ and $TQ_i$ are the successful quantity and the total quantity of subtasks in episode $i$, respectively. $\mathcal{E}$ is the set of evaluation episodes. $\mathbb{E}_{i \in \mathcal{E}}[\,\cdot\,]$ is the mean across $\mathcal{E}$. We consider SR as the primary evaluation metric.

\section{Methodology}
\label{sec:methodology}

\begin{figure*}[t]
\centering
\includegraphics[width=1.0\linewidth]{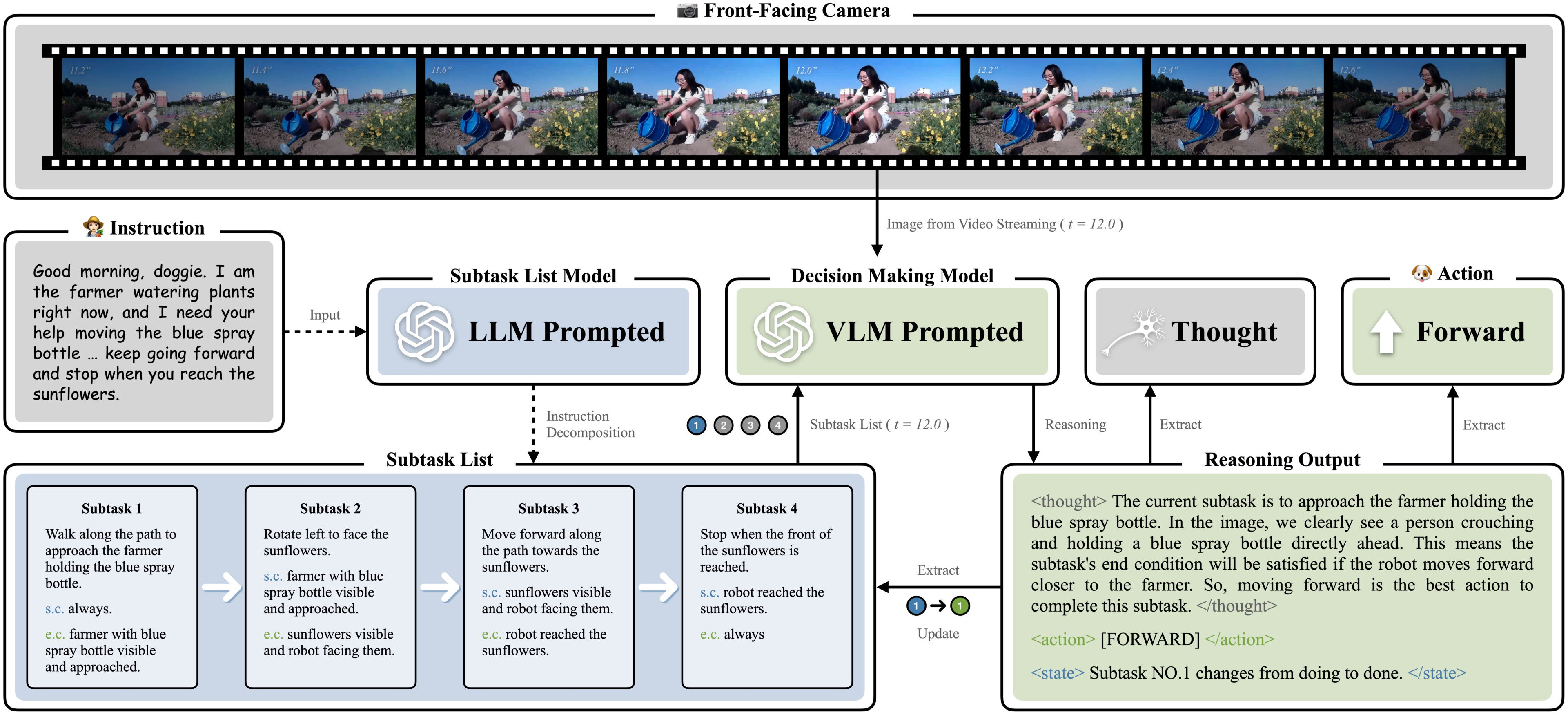}
\caption{AgriVLN method illustration. The instruction decomposition operation drawn by “$\dashrightarrow$” is executed at the beginning for only once, while the decision-making operation drawn by “$\rightarrow$” is executed at every time step. 
The grey, blue and green circles with number $n$ represent the states of \texttt{[pending]}, \texttt{[doing]} and \texttt{[done]} of subtask NO.$n$, respectively. 
}
\label{fig:method}
\end{figure*}

In this section, we present the AgriVLN method, as illustrated in Fig. \ref{fig:method}. First, we present our STL module in Sec. \ref{sec:subtask_list}. Second, we integrate it into the NavGPT \cite{AAAI:NavGPT} backbone to build our AgriVLN method in Sec. \ref{sec:base_model}.

\subsection{Subtask List}
\label{sec:subtask_list} 

\par We prompt a Large Language Model (LLM) to build the instruction decomposer, denoted as $\mathcal{S}\,(\,\cdot\,)$, which decomposes a long instruction $W = \langle w_1, w_2, w_3, \dots, w_L \rangle$ into a list of short subtasks $\langle s_1, s_2, \dots, s_N \rangle$, defined as: 
\begin{equation}
\langle s_1, s_2, \dots, s_N \rangle 
= 
\mathcal{S}\,(W, P_{\mathcal{S}})
\end{equation}
where $N$ is the length of the subtask list. Every subtask $s_i$ consists of five parameters in a structural format, denoted as $s_i = \{ID_i, D_i, SC_i, EC_i, \sigma_i\}$, where $ID$ is the step order, $D$ is the specific description, $SC$ is the condition to start this subtask, $EC$ is the condition to end this subtask, and $\sigma$ is the current state. $P_{\mathcal{S}}$ is the prompt template for $\mathcal{S}\,(\,\cdot\,)$, which is available in our repository, in which we formulate three fundamental principles: 
\begin{itemize}
\item \textbf{Particle Principle}: Any subtask cannot be decomposed into more subdivided subtasks, i.e., $\forall s \in \langle s_1, s_2, \dots, s_N \rangle, \; \neg \exists s', s'', \; \langle s', s'' \rangle = \mathcal{S}\,(s, P_{\mathcal{S}})$.
\item \textbf{Synonymity Principle}: The semantic features of the subtask list must equal to the semantic features of the original instruction, ensuring no information is omitted or added, i.e., $\mathcal{M}\,(W) = \bigcup_{i=1}^{N} \mathcal{M}\,(s_i)$, where $\mathcal{M}\,(\,\cdot\,)$ is the semantic features mapping function.
\item \textbf{Connection Principle}: The starting condition of the next subtask must align with the ending condition of the previous subtask, i.e., $SC_i \simeq EC_{i-1}$. 
\end{itemize}

\subsection{Backbone}
\label{sec:base_model}

To build our AgriVLN method, we use NavGPT \cite{AAAI:NavGPT} as the backbone and integrate our STL module. We prompt a VLM in an end-to-end paradigm to build the decision-maker, denoted as $\mathcal{D}\,(\,\cdot\,)$. At each time step $t$, $\mathcal{D}\,(\,\cdot\,)$ takes both the subtask list $\langle s_1, s_2, \dots, s_N \rangle$ and the camera image $I_t$ as the inputs, to reason the most appropriate low-level action $\hat{a_t}$, defined as: 
\begin{equation}
\hat{a_t}, \Delta \sigma_t, \rho_t 
= 
\mathcal{D}\,(\langle s_1, s_2, \dots, s_N \rangle, I_t, P_{\mathcal{D}})
\end{equation}
where $P_{\mathcal{D}}$ is the prompt template for $\mathcal{D}\,(\,\cdot\,)$, which is available in our repository. $\Delta \sigma_t$ is the transition of state. $\rho_t$ is the thought of reasoning. 

\par We design a total of three candidates for $\sigma$: \texttt{[pending]}, \texttt{[doing]} and \texttt{[done]}. When an instruction $W$ is decomposed into a list of subtasks, all the subtasks are initialized with $\sigma$ = \texttt{[pending]}. For a subtask $s_i$, we formulate the state transition rules as follows: 
\begin{itemize}
\item \texttt{[pending]} $\Rightarrow$ \texttt{[doing]}: If its previous subtask has been completed (i.e., $\sigma_{i-1}$ = \texttt{[done]}) and $\mathcal{D}\,(\,\cdot\,)$ thinks it should start (i.e., $\Delta \sigma_t$ = \textit{“Subtask NO.$i$ changes from pending to doing”}), $\sigma_i$ will be transitioned from \texttt{[pending]} to \texttt{[doing]}.
\item \texttt{[doing]} $\Rightarrow$ \texttt{[done]}: If it is in the state of \texttt{[doing]} (i.e., $\sigma_{i}$ = \texttt{[doing]}) and $\mathcal{D}\,(\,\cdot\,)$ thinks it has been completed (i.e., $\Delta \sigma_t$ = \textit{“Subtask NO.$i$ changes from doing to done”}), $\sigma_i$ will be transitioned from \texttt{[doing]} to \texttt{[done]}. 
\end{itemize}
At each time step $t$, $\mathcal{D}\,(\,\cdot\,)$ attends to the current active subtask $s^{(t)}$ rather than the entire instruction $W$, in which $s^{(t)}$ is defined as: 
\begin{equation}
s^{(t)} 
= 
\begin{cases}
s_i, 
& \text{if } \exists\, i \; \sigma_i = \texttt{[doing]}, \\
\arg\min\limits_{i \in [1,N]}
\left\{ i \mid \sigma_i = \texttt{[pending]} \right\},
& \text{otherwise}.
\end{cases}
\end{equation}
In every episode, $\mathcal{D}\,(\,\cdot\,)$ ends when one of the following conditions happens: 
\begin{quote}
1) $\hat{a_{t'}}$ = $\texttt{[STOP]}$; \\
2) The predicted action sequence $\langle \hat{a_{t'-\tau}}, \hat{a_{t'-\tau+1}}, \dots, \hat{a_{t'}} \rangle$ is deviated to the ground-truth action sequence $\langle a_{t'-\tau}, a_{t'-\tau+1}, \dots, a_{t'} \rangle$; \\
3) $t'$ reaches the max limitation of time step.
\end{quote}
where $t'$ is the stopping time step. $\tau$ is the deviation time threshold, which is set to 4 s as the default value.

\section{Experiments}

\subsection{Experimental Settings}

We implement all the experiments on our A2A benchmark, in which all the open-source LLMs and VLMs are deployed via Ollama \cite{ollama}, and all the closed-source LLMs and VLMs are accessed via APIs. In our AgriVLN method, we adopt GPT-4.1 mini as the default LLM and VLM, and we set their inference temperatures to 1e-4 to ensure the response stability. All the experiments run on a single NVIDIA L20 GPU with 48 G video memory.

\begin{table*}[t]
\captionsetup{skip=8pt}
\caption{Comparison results on the low-complexity portion (subtask $=$ 2), the high-complexity portion (subtask $\geq$ 3), and the whole of our A2A benchmark. \textbf{Bold} and \underline{underline} mark the best and worst scores, respectively.}
\label{tab:comparison_experiment}
\centering
\renewcommand{\arraystretch}{1.3}
\setlength{\tabcolsep}{2pt}
\begin{tabular}{l ccc|ccc|ccc}
\toprule
\multirow{2}{*}{\textbf{Method}} & 
\multicolumn{3}{c}{\textbf{A2A ( low )}} & 
\multicolumn{3}{c}{\textbf{A2A ( high )}} & 
\multicolumn{3}{c}{\textbf{A2A}} \\
\cmidrule(lr){2-4} \cmidrule(lr){5-7} \cmidrule(lr){8-10}
& \textbf{SR} $\uparrow$ & \textbf{NE} $\downarrow$ & \textbf{ISR} $\uparrow$ & 
\textbf{SR} $\uparrow$ & \textbf{NE} $\downarrow$ & \textbf{ISR} $\uparrow$ & 
\textbf{SR} $\uparrow$ & \textbf{NE} $\downarrow$ & \textbf{ISR} $\uparrow$ \\
\midrule
Random    & 0.13 & 7.30          & - & 0.04 & 6.74 & - & 0.09 & 7.03 & - \\
Fixed     & 0.00 & 0.00          & - & 0.06 & 6.32 & - & 0.03 & 3.06 & - \\
\midrule
NavGPT    & 0.51 & \textbf{0.60} & - & 0.14 & 5.01 & - & 0.33 & 2.76 & - \\
SIA-VLN   & 0.52             & 1.46             & 3.27/3.88 & \underline{0.08} & \underline{5.12} & 2.02/4.99 & \underline{0.31} & \underline{3.24} & 2.66/4.42 \\
DILLM-VLN & \underline{0.41} & 1.36             & 4.17/5.08 & 0.32             & 3.90             & 2.59/4.73 & 0.36             & \textbf{2.60}    & 3.40/4.91 \\
\rowcolor{gray!15} 
AgriVLN   & \textbf{0.58}    & \underline{2.32} & 2.01/2.57 & \textbf{0.35}    & \textbf{3.54}    & 1.88/3.23 & \textbf{0.47}    & 2.91             & 1.95/2.89 \\
\midrule
Human     & 0.93 & 0.32 & - & 0.80 & 0.82 & - & 0.87 & 0.57 & - \\
\bottomrule
\end{tabular}
\end{table*}

\subsection{Comparison Experiment}

We compare our AgriVLN method with three existing state-of-the-art VLN methods\footnote{In our reproduction of NavGPT, we remove the depth input to align with our A2A benchmark. In our reproduction of DILLM-VLN, we adopt GLM-4-32b to replace the original GLM-6b for the better fairness.}: NavGPT \cite{AAAI:NavGPT}, SIA-VLN \cite{EMNLP:SIA-VLN} and DILLM-VLN \cite{RAL:DILLM-VLN}. Besides, the methods of Random, Fixed and Human are reproduced as the lower and upper bounds, respectively. 

\par The comparison results are shown in Table \ref{tab:comparison_experiment}. The terrible scores of Random and the excellent scores of Human show the effectiveness and fairness of our A2A benchmark. Compared to the NavGPT baseline, our AgriVLN greatly improves on SR from 0.33 to 0.47 with slight sacrifice on NE from 2.76 m to 2.91 m. When instructions are short and simple, all the methods achieve excellent performances. As the instruction length increases, NavGPT sharply drops on SR from 0.51 to 0.14, while DILLM-VLN and our AgriVLN still keep acceptable performances, from which we suggest the effectiveness of instruction decomposition. Interestingly, SIA-VLN achieves both the second best on the low-complexity portion and the worst on the high-complexity portion. We attribute this gap to its rule-based instruction decomposition: Its lightweight architecture improves the efficiency but its fixed knowledge limits the generalization, which results in the excellent performances on simple tasks but the terrible performances on complex tasks. In contrast, DILLM-VLN and our AgriVLN release the generalization capability of LLM for instruction decomposition, which leads to the relatively consistent performances across simple and complex tasks. 

\par We further compare DILLM-VLN and our AgriVLN in more details. On the low-complexity portion, DILLM-VLN achieves a good NE but the worst SR, while our AgriVLN achieves the best SR but the worst NE. We attribute this distinction to the different reasoning preferences: DILLM-VLN prompts GLM-4-32b simply, so it prefers to aggressively explore more spaces before stopping, which tends to get closer to the target, but tends to collide with it as well. In contrast, our AgriVLN prompts GPT-4.1 mini complexly, so it prefers to conservatively make sure the safety before moving, which tends to arrive within the distance threshold, but tends to stop too early as well. 

\par On the whole of A2A, our AgriVLN achieves an SR of 0.47 and a NE of 2.91 m, showing the state-of-the-art performance in the agricultural VLN domain. Of course, we notice the gap between it and Human, which motivates us to further improve it in the future.

\begin{table}[!t]
\captionsetup{skip=8pt}
\caption{Ablation study results on VLMs. “$_{w/o}$” represents “without”. \textbf{Bold} and \underline{underline} mark the best and worst scores, respectively.}
\label{tab:ablation_experiment_VLM}
\centering
\renewcommand{\arraystretch}{1.3}
\setlength{\tabcolsep}{5pt}
\begin{tabular}{rlccc}
\toprule
\# & \textbf{VLM} & \textbf{SR} $\uparrow$ & \textbf{NE} $\downarrow$ & \textbf{ISR} $\uparrow$ \\ 
\midrule
1 & Qwen2.5-VL-32B                & 0.23             & 4.43             & 1.07/2.64  \\
2 & Qwen2.5-VL-32B $_{w/o}$ STL   & \underline{0.15} & \underline{4.52} & - \\
\midrule
3 & Gemini-1.5 Flash              & 0.24             & 3.51             & 1.79/3.17  \\
4 & Gemini-1.5 Flash $_{w/o}$ STL & 0.20             & 3.41             & - \\
\midrule
5 & Llama-4 Maverick              & 0.33             & 4.22             & 1.55/2.82 \\
6 & Llama-4 Maverick $_{w/o}$ STL & 0.38             & 3.99             & - \\
\midrule
7 & GPT-4.1 mini                  & \textbf{0.47}    & 2.91             & 1.95/2.89 \\
8 & GPT-4.1 mini $_{w/o}$ STL     & 0.33             & \textbf{2.76}    & - \\
\bottomrule
\end{tabular}
\end{table}

\subsection{Ablation Studies}

We implement the ablation studies for our AgriVLN method and STL module on three different aspects: VLMs, subtask quantities, and scene classes. 

\subsubsection{Vision-Language Models}
\label{sec:ablation_study_VLM}
\par We replace GPT-4.1 mini (\#7) with three comparable VLMs, including the open-source representative Qwen2.5-VL-32B (\#1) and the closed-source representatives Gemini-1.5 Flash (\#3) and Llama-4 Maverick (\#5), as shown in Table \ref{tab:ablation_experiment_VLM}. Considering that the GPT-4.1 mini version outperforms the other three VLM versions on both SR and NE, we suggest GPT-4.1 mini as the most matched VLM for AgriVLN.
We further remove STL from all the VLM versions, and observe the significant performance drops across most of them (\#2, \#4, \#8) except for one (\#6), from which we suggest the excellent compatibility of STL on various VLMs.

\begin{table}[!t]
\captionsetup{skip=8pt}
\caption{Ablation study results on subtask quantities. “$_{w/o}$” represents “without”. In every subtask quantity, \textbf{bold} and \underline{underline} mark the better and worse scores when the STL module is integrated, respectively.}
\label{tab:ablation_experiment_subtask_quantity}
\centering
\renewcommand{\arraystretch}{1.3}
\setlength{\tabcolsep}{5pt}
\begin{tabular}{rlcccc}
\toprule
\# & \textbf{Method} & \textbf{Subtask} & \textbf{SR} $\uparrow$ & \textbf{NE} $\downarrow$ & \textbf{ISR} $\uparrow$ \\ 
\midrule
9  &                               & 2        & \textbf{0.58} & \underline{2.32} & 2.01/2.57 \\
10 & AgriVLN                       & 3        & \textbf{0.38} & \textbf{3.04}    & 1.92/3.10 \\
11 &                               & $\geq$ 4 & \textbf{0.14} & \textbf{7.53}    & 1.57/4.29 \\
\midrule
12 &                               & 2        & 0.51 & 0.60 & - \\
13 & AgriVLN $_{w/o}$ STL          & 3        & 0.15 & 4.59 & - \\
14 &                               & $\geq$ 4 & 0.00 & 8.38 & - \\
\bottomrule
\end{tabular}
\end{table}

\begin{table}[!t]
\captionsetup{skip=8pt}
\caption{Ablation study results on scene classes. \textbf{Bold} and \underline{underline} mark the best and worst scores, respectively.}
\label{tab:ablation_experiment_scene}
\centering
\renewcommand{\arraystretch}{1.3}
\setlength{\tabcolsep}{5pt}
\begin{tabular}{rlcccc}
\toprule
\# & \textbf{Scene} & \textbf{Instruction Length} & \textbf{SR} $\uparrow$ & \textbf{NE} $\downarrow$ & \textbf{ISR} $\uparrow$ \\ 
\midrule
15 & Farm       & 40.9 & \textbf{0.66}    & \textbf{1.79}    & 2.13/2.73 \\
16 & Greenhouse & 42.1 & 0.35             & 2.56             & 1.88/3.07 \\
17 & Forest     & 48.0 & \underline{0.33} & 3.68             & 1.78/2.86 \\
18 & Mountain   & 40.9 & 0.50             & 2.84             & 1.94/2.88 \\
19 & Garden     & 53.3 & 0.49             & \underline{3.77} & 2.00/3.07 \\
20 & Village    & 55.9 & 0.47             & 2.92             & 1.93/2.73 \\
\bottomrule
\end{tabular}
\end{table}

\subsubsection{Subtask Quantities}

We remove STL from AgriVLN, as shown in Table \ref{tab:ablation_experiment_subtask_quantity}. When the subtask quantity is 2 (\#9, \#12), the task complexity is relatively simple, so STL shows no significant effects. As the subtask quantity increases, the overall task complexity gradually escalates, so STL starts to show effects. When the subtask quantity is increased to 4 (\#11, \#14), AgriVLN without STL achieves an SR of 0.00, i.e., not even one episode can be succeeded. However, AgriVLN still keeps the acceptable performance with an SR of 0.14 and a NE of 7.53 m, from which we suggest that STL can effectively decompose a complex instruction into a list of simple subtasks, thereby assisting the decision-maker to focus on one task at any time.

\subsubsection{Scene Classes}

We respectively summarize the performances of AgriVLN on different scene classes, as shown in Table \ref{tab:ablation_experiment_scene}. The average instruction lengths are relatively consistent, from which we suggest no significant differences in the intrinsic difficulties of the instructions themselves. However, AgriVLN shows substantial variations across different scene classes, such as the 33-point gap on SR between farms (\#15) and forests (\#17) and the 1.98-m gap on NE between farms (\#15) and gardens (\#19). A possible reason is that the subtle visual distinctions across different scene classes - such as background clutters, obstacle densities, and light conditions - pose varying levels of visual perception challenges to the decision-maker. On the one hand, these gaps validates the value to collect data from diversified scene classes, which supports our A2A to be a comprehensive benchmark for evaluating agricultural VLN methods. On the other hand, these gaps highlights the need to further improve the robustness against challenging scene classes in the future, such as forests and gardens.

\begin{figure*}[t]
\centering
\includegraphics[width=1.0\linewidth]{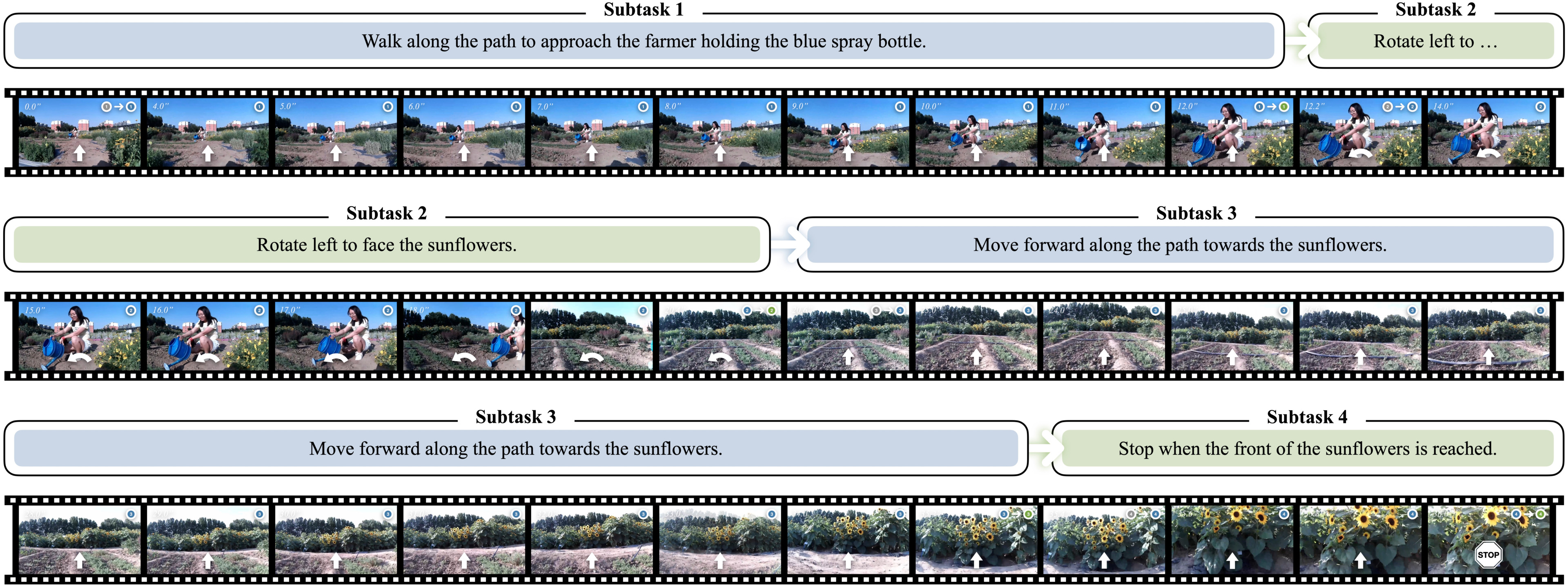}
\caption{Case study illustration. The grey, blue and green circles with number $n$ represent the states of \texttt{[pending]}, \texttt{[doing]} and \texttt{[done]} of subtask NO.$n$, respectively. (Please zoom in for a better observation)}
\label{fig:qualitative_experiment}
\end{figure*}

\subsection{Case Study}

We implement the case study on the representative episode, as illustrated in Fig. \ref{fig:qualitative_experiment}. In this episode, the target position is a cluster of sunflowers. After receiving the instruction from user, our AgriVLN method calls the STL module to decompose it into a list of four subtasks, then executes each subtask one by one, thereby reasoning a sequence of low-level actions to navigate the agricultural robot to move $\texttt{[FORWARD]}$ to reach the user, $\texttt{[LEFT ROTATE]}$ to face the target sunflowers, move $\texttt{[FORWARD]}$ to approach them, and $\texttt{[STOP]}$ when reaching them. From this successful case, we suggest that AgriVLN can effectively address the problems on limited mobility and poor adaptability for agricultural robots.

\section{Conclusion}

In this paper, we build the A2A benchmark, pioneeringly extending VLN for agricultural robots. We evaluate several existing state-of-the-art VLN methods on it, then we observe that they perform well on short instructions while perform poorly on long instructions. To address this issue, we design the STL module, which decomposes a complex instruction into a list of simple subtasks, thereby assisting the decision-maker to focus on one task at any time. We integrate it into the NavGPT backbone to propose our AgriVLN method, which achieves an SR of 0.47 and a NE of 2.91 m, showing the state-of-the-art performance in the agricultural VLN domain.

\par During the experiments, we also find a main limitation: The time cost of the VLM-based decision-making on a single time step is about 0.2 s $\sim$ 0.3 s, from which we suggest a further improvement on the real-time capability \cite{CVPR:Frequency,JRTIP}.

\par In the future, in addition to the improvement on the existing limitation, we plan to explore would agricultural VLN methods perform better if we could polish instructions to make them easier to understand \cite{AAAI:SepPrune}, and design a new translator \cite{ICIC:T-araVLN} to make it come true.

\begin{credits}

\subsubsection{\ackname} 
This work is supported by the Sichuan Chengdu Modern Agricultural Industry Research Institute of China Agricultural University: Provincial and Municipal Agricultural Subsidy Funded Project; the Natural Science Foundation of Sichuan Province (2024NSFSC0389); and the Provincial and Municipal Agricultural Subsidy Special Funds for the Construction of CAU–SCCD Advanced Agricultural and Industrial Institute. This work is also supported by Xiaobei Zhao. Thanks to Beijing, Tbilisi, Baku, Kunming, Pokhara and Moalboal for the impressive traveling experiences, giving us a chilled vibe for experiment and writing. Thanks to Yuanquan Xu, the inspiration to us.


\subsubsection{\discintname}
The authors have no competing interests to declare that are relevant to the content of this article. 

\end{credits}

\bibliographystyle{splncs04}
\bibliography{iconip26}

\end{document}